\documentclass[11pt]{article}
\usepackage{deauthor,times,graphicx}

\usepackage{amsthm}
\usepackage{graphicx}
\usepackage{subcaption}
\usepackage{xcolor}
\usepackage{floatrow}
\newfloatcommand{capbtabbox}{table}[][\FBwidth]

\graphicspath{{figs/}}

\newcommand{\slicetuner}{Slice Tuner}
\newcommand{\frtrain}{FR-Train}
\newcommand{\fb}{FairBatch}
\newcommand{\mc}{MLClean}
\newcommand{\slicefinder}{Slice Finder}

\definecolor{blue(ryb)}{rgb}{0.01, 0.28, 0.9}

\begin{document}
\title{Responsible AI Challenges in End-to-end Machine Learning}


\author{Steven Euijong Whang, Ki Hyun Tae, Yuji Roh, Geon Heo \\
KAIST\\
{\small \{swhang, kihyun.tae, yuji.roh, geon.heo\}@kaist.ac.kr}}

\maketitle

\begin{abstract}

Responsible AI is becoming critical as AI is widely used in our everyday lives. Many companies that deploy AI publicly state that when training a model, we not only need to improve its accuracy, but also need to guarantee that the model does not discriminate against users (fairness), is resilient to noisy or poisoned data (robustness), is explainable, and more. In addition, these objectives are not only relevant to model training, but to all steps of end-to-end machine learning, which include data collection, data cleaning and validation, model training, model evaluation, and model management and serving. Finally, responsible AI is conceptually challenging, and supporting all the objectives must be as easy as possible. We thus propose three key research directions towards this vision -- {\em depth}, {\em breadth}, and {\em usability} -- to measure progress and introduce our ongoing research. First, responsible AI must be deeply supported where multiple objectives like fairness and robust must be handled together. To this end, we propose \frtrain{}, a holistic framework for fair and robust model training in the presence of data bias and poisoning. Second, responsible AI must be broadly supported, preferably in all steps of machine learning. Currently we focus on the data pre-processing steps and propose \slicetuner{}, a selective data acquisition framework for training fair and accurate models, and \mc{}, a data cleaning framework that also improves fairness and robustness. Finally, responsible AI must be usable where the techniques must be easy to deploy and actionable. We propose \fb{}, a batch selection approach for fairness that is effective and simple to use, and \slicefinder{}, a model evaluation tool that automatically finds problematic slices. We believe we scratched the surface of responsible AI for end-to-end machine learning and suggest research challenges moving forward.

\end{abstract}

\section{Introduction}

Responsible AI is becoming critical as machine learning becomes widespread in our everyday lives. Companies including Google~\cite{google}, Microsoft~\cite{microsoft}, and IBM~\cite{ibm} publicly state that AI not only needs to be accurate, but also used and developed, evaluated, and monitored for trust. Although there is no universally agreed notion for responsible AI, the major objectives include fairness, robustness, explainability, transparency, and accountability.

The usual starting point is to support responsible AI only in model training, but this is not sufficient. For example, if the training data is biased towards a specific population, there is a fundamental limit into how much the trained model can avoid being biased as well even using the best fair training algorithms. Instead, we may need to address the root cause starting from data collection where we need to construct an unbiased dataset.

We would thus like to support responsible AI in all steps of end-to-end machine learning~\cite{DBLP:conf/kdd/BaylorBCFFHHIJK17,DBLP:journals/debu/ZahariaCD0HKMNO18}. Before model training, the key steps are data collection, data cleaning, and validation. After model training, there are model evaluation, and model management and serving. In addition, since supporting all the responsible AI objectives is already conceptually challenging, it is important to make these techniques easy to use as well.

To this end, we propose three research directions -- {\em depth}, {\em breadth}, and {\em usability} -- and present our contributions. First, we need to deeply support responsible AI where multiple objectives are addressed together. We present FR-Train~\cite{DBLP:conf/icml/RohLWS20}, the first holistic framework for fair and robust model training.
Second, we need to broadly support responsible AI in all machine learning steps. We present two systems that focus on data pre-processing: \slicetuner{}~\cite{DBLP:journals/corr/abs-2003-04549} is a selective data acquisition framework for fair and accurate models, and \mc{}~\cite{DBLP:conf/sigmod/TaeROKW19} is a data cleaning framework that also improves fairness and robustness.
Third, we need responsible AI to be usable and actionable. We present two systems: \fb{}~\cite{fairbatch} is an easy-to-deploy batch selection technique for model training that improves fairness, and \slicefinder{}~\cite{DBLP:conf/icde/ChungKPTW19,DBLP:journals/tkde/ChungKPTW20} automatically evaluates a model by finding problematic slices where it underperforms.
Our work only scratches the surface of responsible AI for end-to-end machine learning, and we believe that setting the three research directions is useful to measure progress.

We introduce the responsible AI research landscape in Section~\ref{sec:landscape}. We then discuss our systems for depth, breadth, and usability in Sections~\ref{sec:depth}, \ref{sec:breadth}, and \ref{sec:usability}, respectively. Finally, we suggest open challenges in Section~\ref{sec:future}.

\section{Responsible AI Research Landscape}
\label{sec:landscape}

We provide a brief history of responsible AI and discuss the research landscape. Responsible AI is also known as Trustworthy AI and has recently been promoted by Google~\cite{google}, Microsoft~\cite{microsoft}, and IBM~\cite{ibm} among others as a critical issue when using AI in practice. The key objectives include fairness, robustness, explainability, transparency, and accountability. Among the objectives, we focus on fairness and robustness because they are both closely related to the training data. The other objectives are also important, but currently outside our scope.

Fairness is the problem of not discriminating against users and has gained explosive interest in the past decade~\cite{barocas-hardt-narayanan, DBLP:conf/pods/Venkatasubramanian19}. An article that popularized fairness was the 2016 ProPublica report~\cite{machinebias} on the COMPAS software, which is used in US courts to predict a defendant's recidivism (reoffending) rate. COMPAS is convenient, but is known to overestimate black people's recidivism risk compared to white people. Recently, various unfairness mitigation techniques~\cite{aif360-oct-2018} have been proposed and can be categorized as pre-processing, in-processing, or post-processing depending on whether the techniques are applied before, during, or after model training, respectively.

Robustness is the problem of preventing or coping with adversarial attacks. In particular, model training against data poisoning has been heavily studied in the past decade~\cite{DBLP:conf/sp/CretuSLSK08,DBLP:journals/corr/abs-2007-08199}. Nowadays datasets are easier to publish using tools like Kaggle and Google Dataset Search~\cite{DBLP:conf/semweb/BenjellounCN20}, which means that it is easier to disseminate poisoned data as well. The data can then be harvested by Web crawlers of unsuspecting victims and used for model training. While the basic poisoning attacks involve simple labeling flipping (e.g., change a positive label to be negative), recent poisoning attacks are becoming increasingly sophisticated.
The possible defenses include sanitizing the data before model training or making the model training accurate despite the poisoning.

In practice, machine learning is not just about model training, but involves multiple steps as demonstrated by end-to-end systems like TensorFlow Extended (TFX)~\cite{DBLP:conf/kdd/BaylorBCFFHHIJK17} and MLFlow~\cite{DBLP:journals/debu/ZahariaCD0HKMNO18}: data collection, data cleaning and validation, model training, model evaluation, and model management and serving. Hence, responsible AI is not just a model training issue, but relevant to all of the above steps. The data management community has recently been addressing the data aspect of responsible AI in end-to-end machine learning~\cite{DBLP:conf/sigmod/Polyzotis0WZ17,DBLP:conf/sigmod/Kumar0017,DBLP:journals/sigmod/PolyzotisRWZ18,breck2019data,DBLP:journals/tkde/RohHW19,DBLP:journals/pvldb/AsudehJ20a,DBLP:journals/pvldb/Whang020}. 

\begin{figure}[t]
  \centering
  \includegraphics[width=0.8\textwidth]{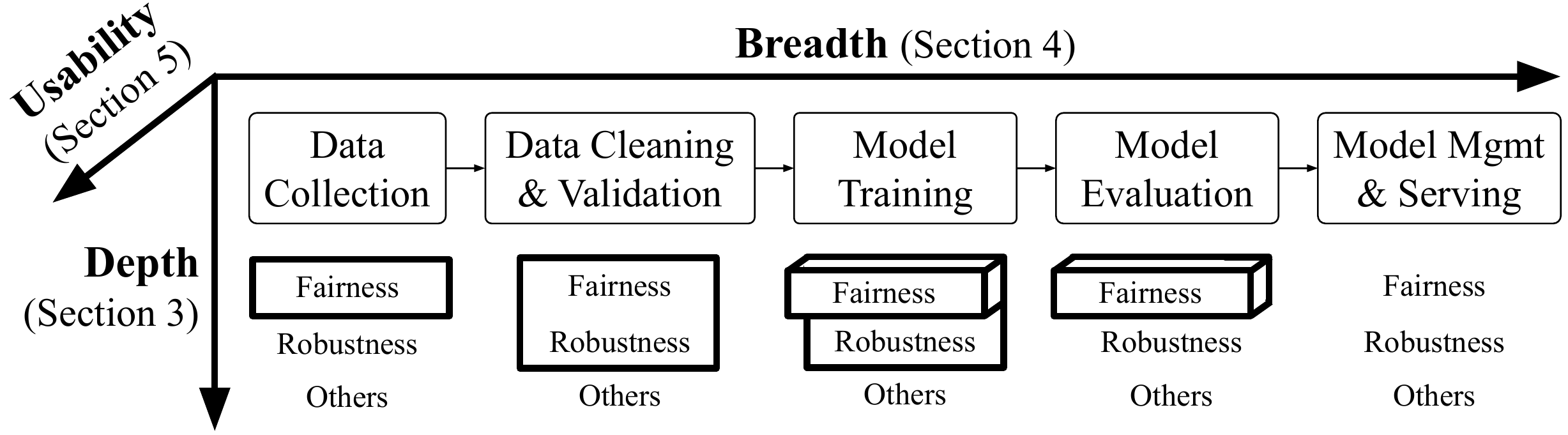}
  \caption{The three research directions -- depth, breadth, and usability -- for fully supporting the responsible AI objectives (fairness, robustness, and others) in addition to accuracy in end-to-end machine learning. The highlighted parts show our contributions:  \slicetuner{}~\cite{DBLP:journals/corr/abs-2003-04549} addresses fairness in data collection; \mc{}~\cite{DBLP:conf/sigmod/TaeROKW19} addresses fairness and robustness in data cleaning; \frtrain{}~\cite{DBLP:conf/icml/RohLWS20} addresses fairness and robustness in model training; \fb{}~\cite{fairbatch} addresses usability for fairness in model training; and \slicefinder{}~\cite{DBLP:conf/icde/ChungKPTW19,DBLP:journals/tkde/ChungKPTW20} addresses usability for fairness in model evaluation.}
  \label{fig:responsibleai}
\end{figure}

The current research landscape naturally leads to the three key research directions we propose -- {\em depth}, {\em breadth}, and {\em usability} -- as shown in  Figure~\ref{fig:responsibleai}. First, it is important to support many responsible AI objectives at each step. Second, we need to broadly support responsible AI in as many steps as possible, from data collection to model serving. Third, we need these techniques to be usable and actionable by machine learning users. 
We highlight the responsible AI objectives in Figure~\ref{fig:responsibleai} where we propose solutions.

\section{Deep Responsible AI}
\label{sec:depth}

We discuss deeply supporting responsible AI, which means that we would like to address multiple objectives together. We re-emphasize that each objective is currently being heavily studied. For model fairness, there is an extensive literature in the machine learning and fairness communities on mitigating unfairness before, during, or after model training~\cite{barocas-hardt-narayanan,DBLP:conf/pods/Venkatasubramanian19,aif360-oct-2018}. For model robustness, both the machine learning and security communities are proposing various data sanitization and robust training techniques~\cite{DBLP:journals/corr/abs-1811-00741,DBLP:journals/corr/abs-2007-08199}. However, we believe that responsible AI requires both fairness and robustness instead of just one. In addition, addressing one objective at a time is not ideal as we discuss later. Fairness and robustness are also closely related because their problems originate from the training data: biased data causes unfairness while poisoned data decreases model accuracy. This motivation leads us to propose \frtrain{}~\cite{DBLP:conf/icml/RohLWS20}, the first holistic framework for fair and robust training.

Fairness is a subjective notion, and many definitions have been proposed~\cite{DBLP:conf/pods/Venkatasubramanian19} where they can be categorized depending on what information is used: the classifier, the sensitive attribute (e.g., race or gender), and training labels. For example, individual fairness only uses the classifier and means that similar individuals must have similar predictions. Demographic parity~\cite{DBLP:conf/kdd/FeldmanFMSV15} (or disparate impact) uses the classifier and the protected attribute and means that different sensitive groups (e.g., black and white populations) have similar positive prediction rates. That is, $P(\hat{Y}=1|Z=0) \approx P(\hat{Y}=1|Z=1)$ where $\hat{Y}$ is a prediction and $Z$ is a binary sensitive attribute. Equalized odds~\cite{DBLP:conf/nips/HardtPNS16} uses all three pieces of information and is similar to demographic parity, except that the probabilities are conditioned on the label. That is, $P(\hat{Y}=1|Z=0,Y=l) \approx P(\hat{Y}=1|Z=1,Y=l)$ where $Y$ is the label. In this section, we use demographic parity and measure it using the formula $DP := \min \left( \frac{P(\hat{Y}=1|Z=0)}{P(\hat{Y}=1|Z=1)}, \frac{P(\hat{Y}=1|Z=1)}{P(\hat{Y}=1|Z=0)} \right)$ where a higher value close to 1 means better fairness.

\begin{figure}[t]
\begin{subfigure}[t]{0.49\textwidth}
\centering
\includegraphics[scale=0.5]{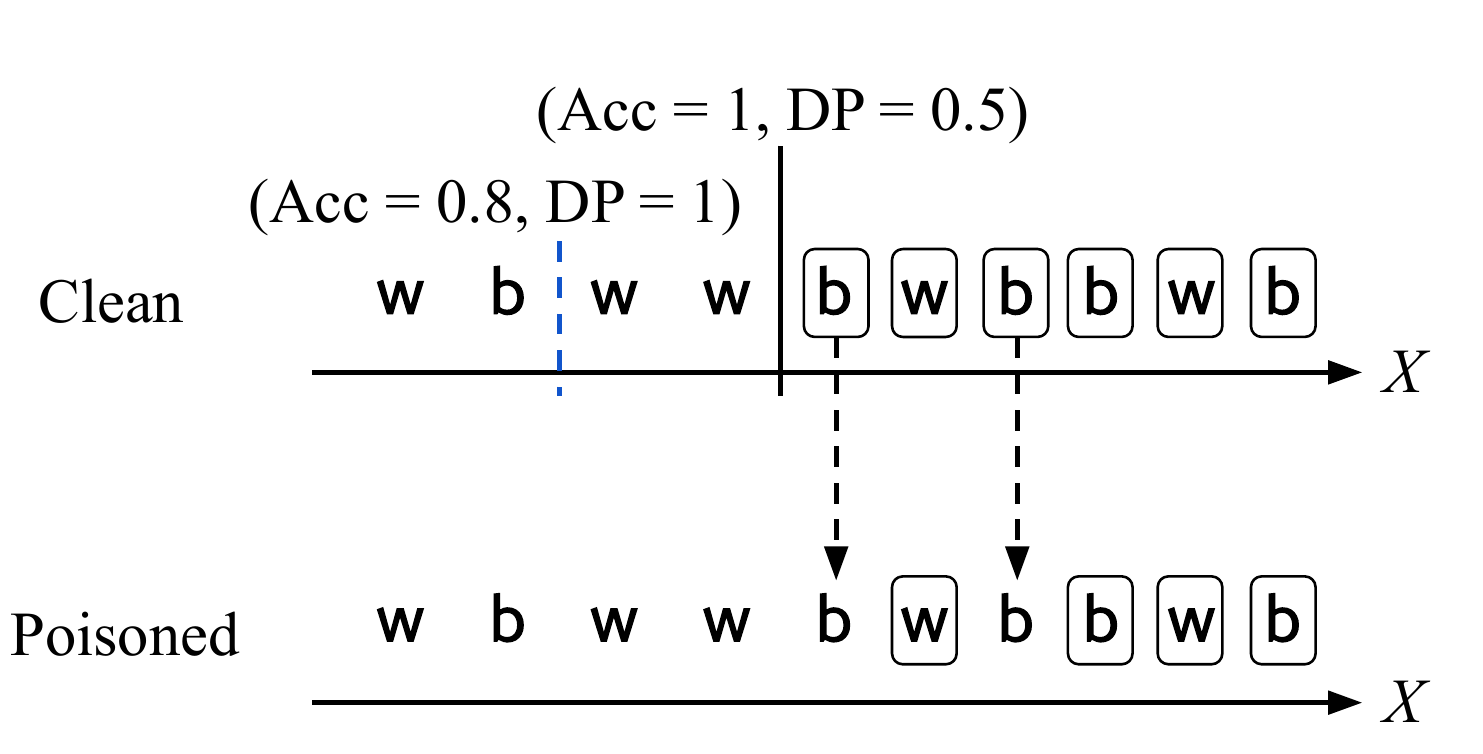}
\caption{}
\label{fig:tradeoff1}
\end{subfigure}
\begin{subfigure}[t]{0.49\textwidth}
\includegraphics[scale=0.5]{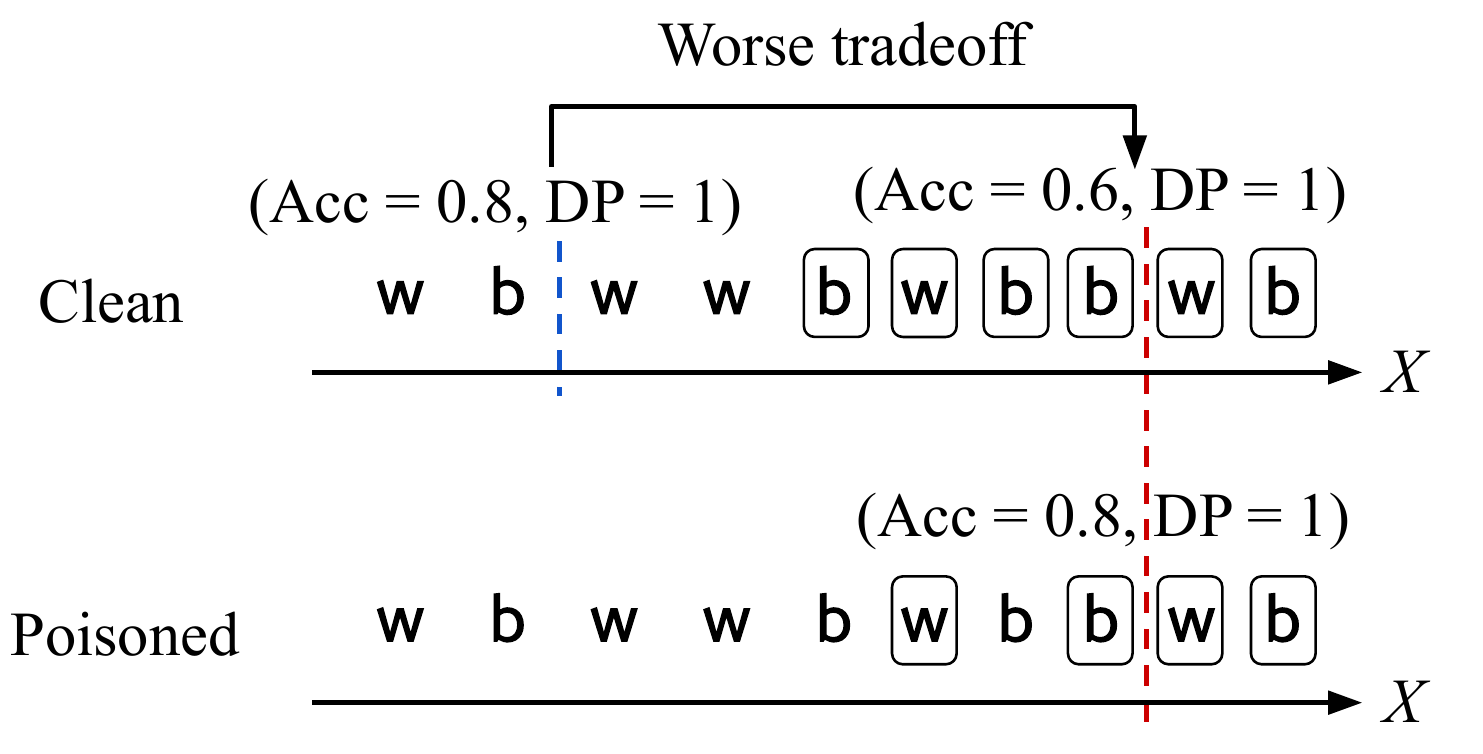}
\caption{}
\label{fig:tradeoff2}
\end{subfigure}
  \caption{(a) Accurate (black solid line) and fair (blue dotted line) classifiers on clean data followed by data poisoning. (b) A fair classifier trained on poisoned data (red dotted line) is evaluated on clean data, showing a worse accuracy-fairness tradeoff than the fair classifier trained on clean data.}
  \label{fig:tradeoff}
\end{figure}

We now explain why addressing fairness and robustness together is important using a concrete example. In Figure~\ref{fig:tradeoff1}, suppose there are two sensitive groups black and white, and that there are ten people of two races: white (denoted as `w') and black (denoted as `b'). Let us assume the boxes indicates positive labels and that we want to train a threshold classifier that divides the individuals using a single feature $X$ where those on the left have negative predictions (e.g., do not reoffend) while those on the right have positive predictions. On clean data, a vanilla classifier can obtain perfect accuracy by dividing between the fourth and fifth individuals (Figure~\ref{fig:tradeoff1} solid line classifier). However, the demographic parity $DP$ is not perfect where $P(\hat{Y}=1|Z=w) = \frac{2}{5} = 0.4$, $P(\hat{Y}=1|Z=b) = \frac{4}{5} = 0.8$, and $DP := \min \left( \frac{0.4}{0.8}, \frac{0.8}{0.4} \right) = 0.5$. Suppose a fair classifier maximizes accuracy with perfect $DP$. One can find such a classifier by dividing between the second and third individuals (Figure~\ref{fig:tradeoff1} blue dotted line classifier). While $DP = 1$, the accuracy is 0.8 because two white people are now misclassified. 

Now suppose we poison the clean data by using the standard method of flipping labels~\cite{DBLP:conf/pkdd/PaudiceML18}. On the bottom of Figure~\ref{fig:tradeoff1}, the fifth and seventh individuals are now incorrectly labeled as negative. There are three ways to handle the poisoned data: (1) do nothing and perform fair training only as usual, (2) take a two-step approach and perform data sanitization followed by fair training using existing methods, and (3) take a holistic approach for fair and robust training. Let us first see what happens if we take the first approach. We can train a fair classifier on the poisoned data with perfect $DP$ by dividing between the eighth and ninth individuals (bottom of Figure~\ref{fig:tradeoff2}, red dotted line classifier). In that case, we will have perfect $DP$, but an accuracy of 0.8 on poisoned data. However, if this classifier is deployed in the real world, it will effectively be used on clean data. This scenario is plausible for any application that serves real customers. However, simply using the same classifier on clean data results in a worse tradeoff of fairness and accuracy where $DP$ remains the same, but the accuracy reduces to 0.6. Hence,  ignoring poisoning may lead to strictly worse accuracy and fairness results. In reference~\cite{DBLP:conf/icml/RohLWS20}, we also empirically show that the two-step solution is ineffective. The intuition is that an existing fairness-only or robustness-only technique cannot easily distinguish data poisoning from bias in the data and ends up removing all or none of the problematic data.

\begin{figure}[t]
  \centering
  \includegraphics[width=0.78\textwidth]{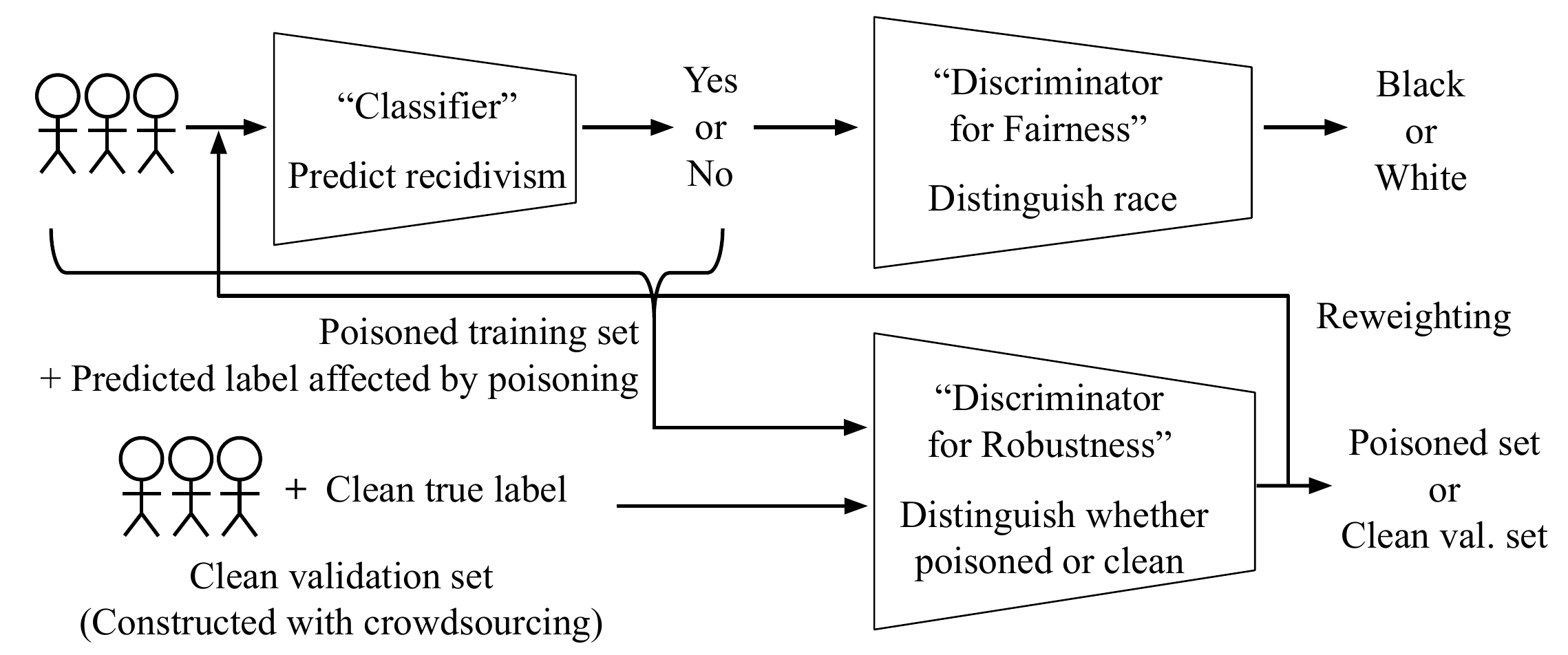}
  \caption{The \frtrain{} architecture and how it can be used for recidivism prediction.}
  \label{fig:frtrain}
\end{figure}

We thus propose \frtrain{} to take a holistic approach for fair and robust training. Figure~\ref{fig:frtrain} shows the architecture of \frtrain{}. On the top, there is a classifier (e.g., predicts recidivism) that competes with a discriminator for fairness that predicts the sensitive attribute (e.g., the race) based on the predictions. This adversarial training is similar to Adversarial Debiasing~\cite{DBLP:conf/aies/ZhangLM18}, a state-of-the-art fairness-only training algorithm. The below part is the novel addition where there is a discriminator for robustness that distinguishes the possibly-poisoned training set with a validation set that is known to be clean. The clean validation set is small and can be constructed using crowdsourcing and conventional quality control techniques including majority voting. Hence, the classifier needs to be both fair and robust to compete with the two discriminators. Finally, the predictions of the robustness discriminator are used to reweight training set examples where cleaner examples get higher weights. Initially, these weights are not useful because the robustness discriminator is not accurate. However, as the training progresses, the discriminator becomes accurate, and the weights are used by the classifier.

In reference~\cite{DBLP:conf/icml/RohLWS20}, we present a mutual information-based interpretation of \frtrain{}'s architecture. To give an intuition, perfect fairness means that the mutual information between the model's prediction and the sensitive attribute is 0. Similarly, satisfying robustness can be expressed using mutual information. In this case, perfect robustness means that the poisoned data distribution is indistinguishable from the clean data distribution (i.e., validation set). \frtrain{} minimizes both of the mutual information values and the classifier loss. We perform experiments on synthetic and real datasets and train a classifier on poisoned data and evaluate it on clean data. As a result, \frtrain{} is the only approach that achieves both high accuracy and fairness while the other baselines either have poor fairness or accuracy.


\section{Broad Responsible AI}
\label{sec:breadth}

In addition to supporting responsible AI in model training, we would also like to broadly support it across many steps in end-to-end machine learning. While most of the fairness and robustness literature focus on model training, there needs to be more focus on other machine learning steps as well. Recently, FairPrep~\cite{DBLP:conf/edbt/SchelterHKS20} was proposed to support fairness in all steps of data pre-processing before model training. Also for an extensive coverage of data collection and quality techniques for machine learning, please refer to a survey~\cite{DBLP:journals/tkde/RohHW19} and tutorial~\cite{DBLP:journals/pvldb/Whang020}.
Here we also focus on data pre-processing and present two contributions: \slicetuner{}~\cite{DBLP:journals/corr/abs-2003-04549} is a selective data acquisition framework for maximizing fairness and accuracy, and \mc{}~\cite{DBLP:conf/sigmod/TaeROKW19} is a data cleaning framework for addressing both fairness and robustness in addition to accuracy.

\subsection{Selective Data Acquisition for Fair and Accurate Models}
\label{sec:slicetuner}

As machine learning is used in various applications, one of the critical bottlenecks is acquiring enough data so that the trained model is both accurate and fair. Nowadays, there are many ways to acquire data including dataset discovery, crowdsourcing, and simulator-based data generation. Data acquisition is not the same as active learning, which labels existing data. Instead, our focus is on acquiring new data along with its labels. 

However, blindly acquiring data is not the right approach. Let us first divide the data into subsets called slices. Suppose that the slices are customer purchases by various regions: America, Europe, APAC, and so on. Among them, if we already have enough America data, acquiring more America data is not only unhelpful, but may also bias the data and have a negative effect on the model accuracy on the other slices.

Instead, we want to acquire possibly different amounts of data per slice in order to maximize accuracy and fairness. To measure accuracy, we use loss functions like logistic loss. For fairness, we use equalized error rates~\cite{DBLP:conf/pods/Venkatasubramanian19}, which states that the losses of slices must be similar. This notion of fairness is important to any application that should not discriminate its customers by service quality. A waterfilling approach is a good start where we simply acquire data so that the slices have similar sizes. However, this approach is not optimal because some slices may need more data to obtain the same model loss as other slices.

\begin{figure}[t]
\begin{subfigure}[t]{0.48\textwidth}
\centering
\includegraphics[scale=0.3]{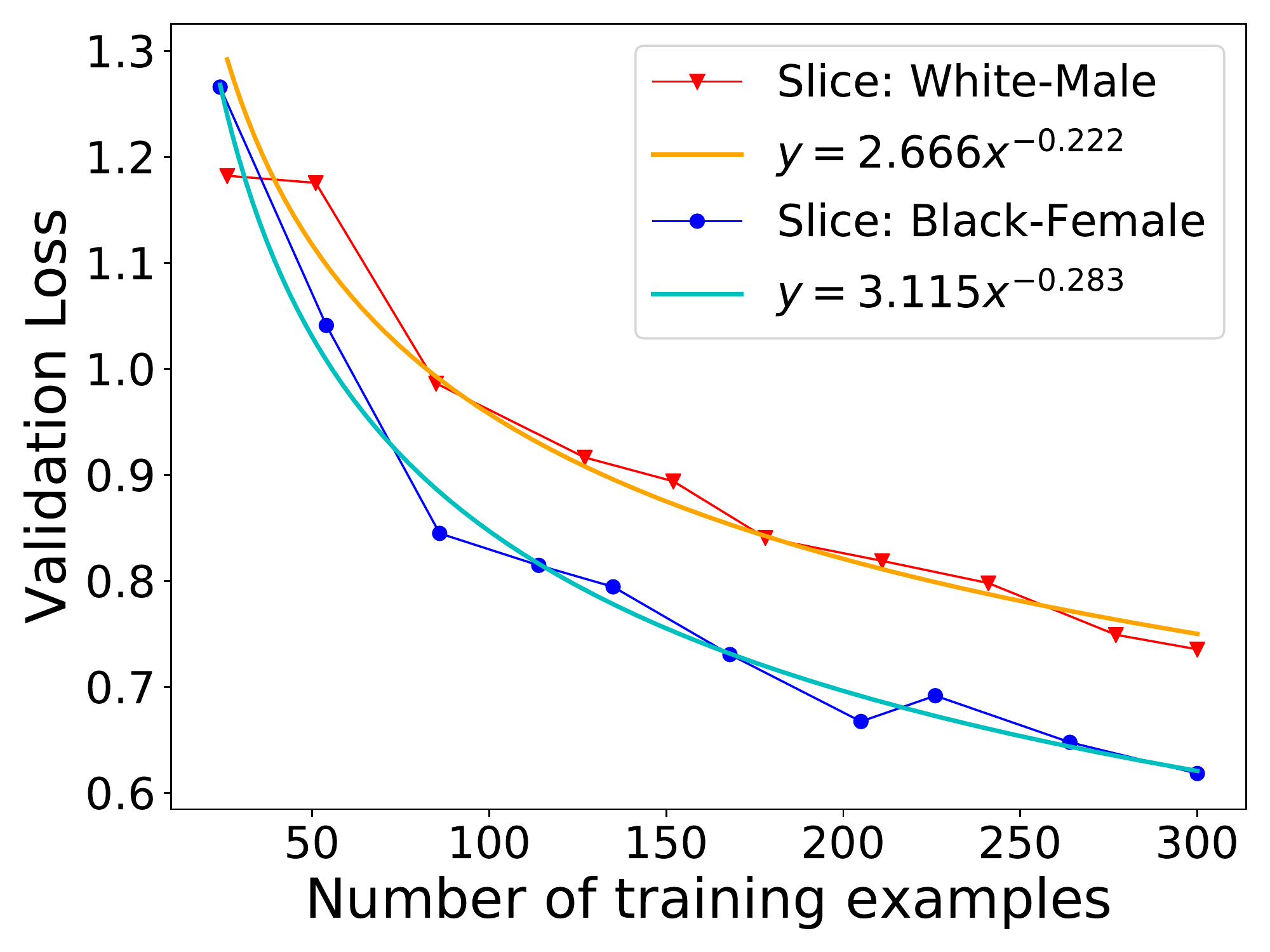}
\caption{}
\label{fig:learningcurve}
\end{subfigure}
\begin{subfigure}[t]{0.51\textwidth}
\includegraphics[scale=0.65]{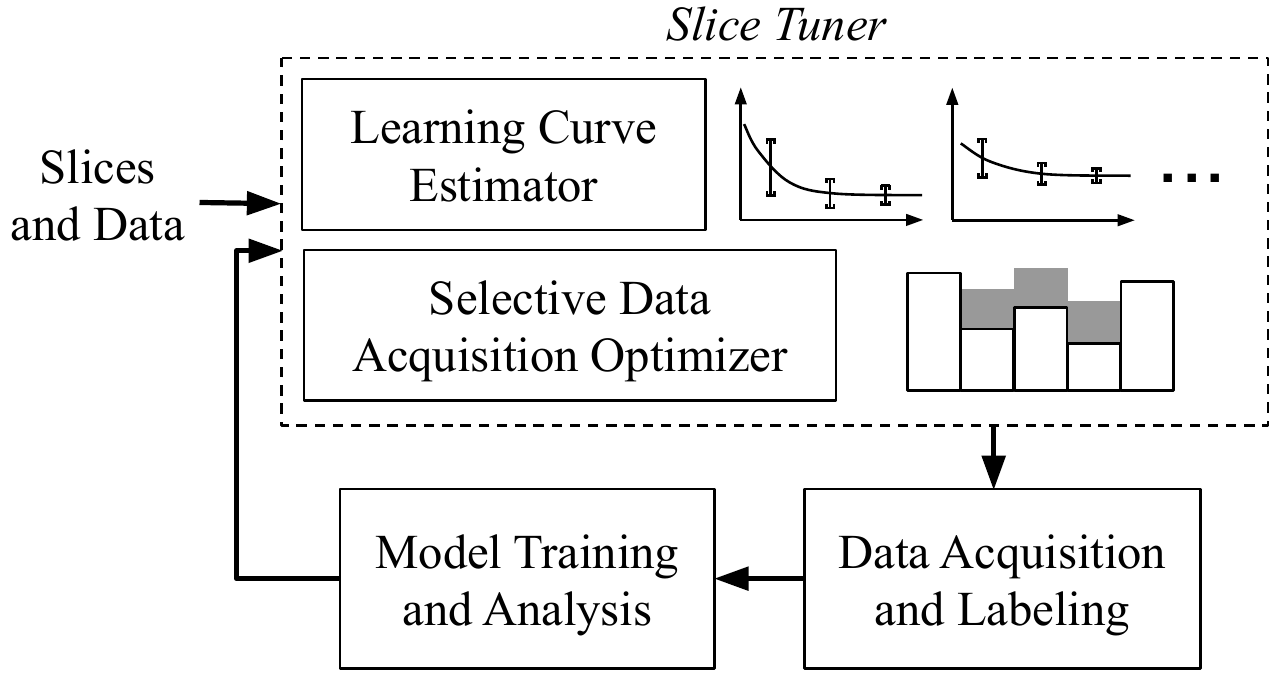}
\caption{}
\label{fig:slicetuner}
\end{subfigure}
  \caption{(a) Learning curves on two slices of the UTKFace dataset~\cite{zhifei2017cvpr}. (b) Slice Tuner architecture.}
  \label{fig:}
\end{figure}

Our key approach is to generate for each slice a {\em learning curve}, which estimates the model loss on that slice given more labeled data. Multiple studies~\cite{baidu2017deep,DBLP:conf/ijcai/DomhanSH15} show that a learning curve is best fit using a power-law function. Figure~\ref{fig:learningcurve} shows two actual learning curves generated on two race-gender slices of a real dataset called UTKFace~\cite{zhifei2017cvpr}. We can use these learning curves to estimate how much data must be acquired per slice.

Assuming that the learning curves are perfectly reliable (we discuss how to deal with unreliable curves later), we can determine the amounts of data to acquire to minimize the total loss and unfairness of slices by solving the following convex optimization problem:
\[\min \sum_{i=1}^n b_i(|s_i| + d_i)^{-a_i} + \lambda \sum_{i=1}^n \max \left\{0, \frac{b_i(|s_i| + d_i)^{-a_i}}{A} - 1 \right\} \textrm{\ subject to\ } \sum_{i=1}^n C(s_i) \times d_i = B
\]
where $\{s_i\}_{i=1}^n$ are the slices, $\{d_i\}_{i=1}^n$ are the amounts of data to acquire, $A$ is the average loss of slices, $C(s_i)$ is the cost function for acquiring an example for $s_i$, and $B$ is a cost budget. The first term in the objective function minimizes the total loss while the second term minimizes the unfairness by penalizing slices that have higher-than-average losses. The two terms are balanced using $\lambda$. By acquiring more data for slices with higher losses, we eventually satisfy equalized error rates. \slicetuner{}'s architecture is shown in Figure~\ref{fig:slicetuner} where we perform selective data acquisition on input slices. The runtime bottleneck is the time to actually acquire data.

\begin{figure}[h]
\begin{subfigure}[t]{0.51\textwidth}
\centering
\includegraphics[scale=0.23]{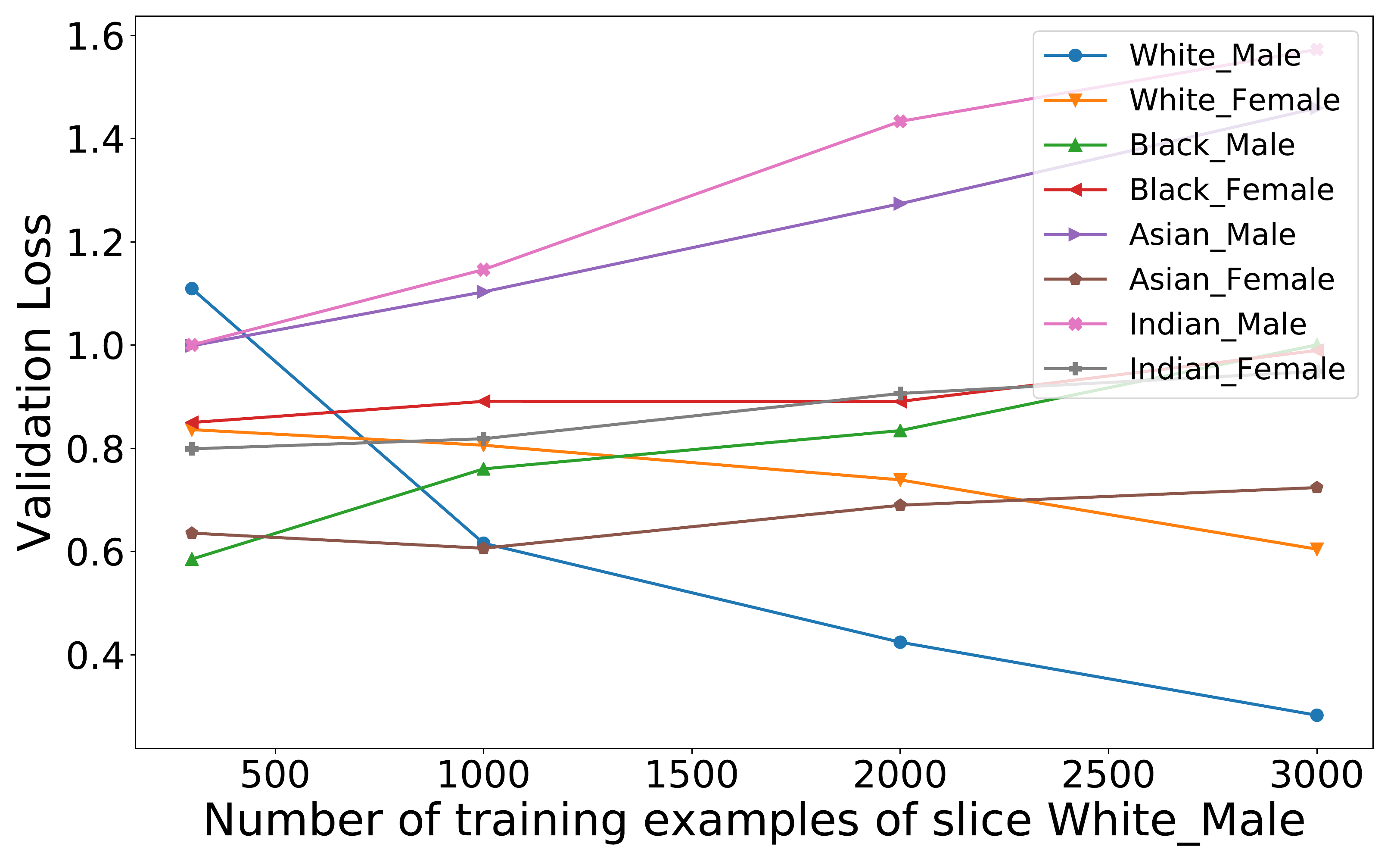}
\caption{}
\label{fig:utkfaceinfluence}
\end{subfigure}
\begin{subfigure}[t]{0.48\textwidth}
\centering
\includegraphics[scale=0.54]{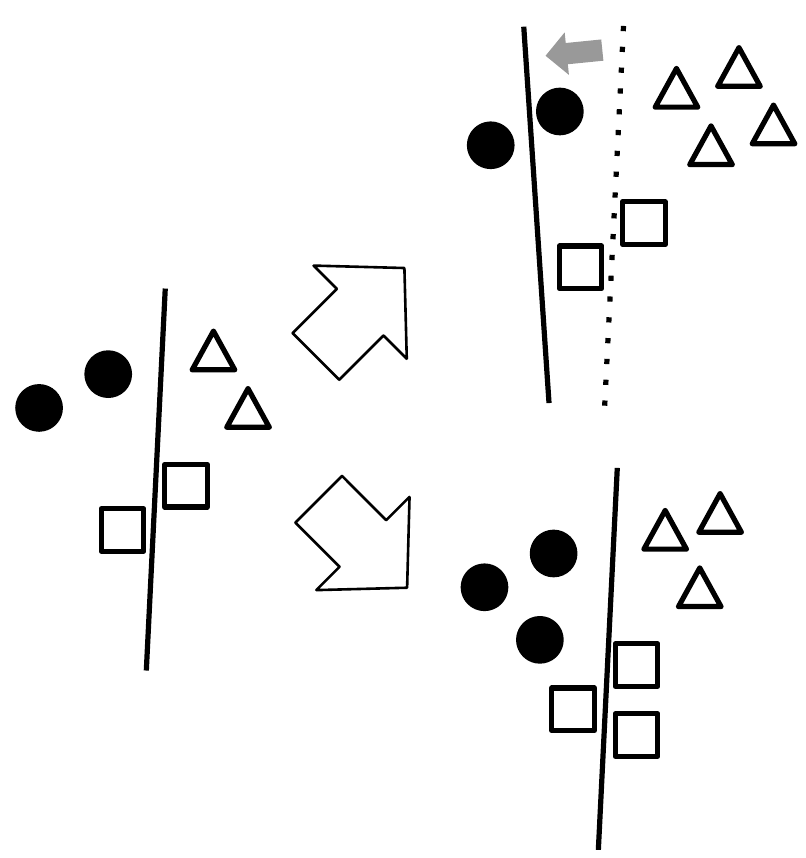}
\caption{}
\label{fig:influenceintuition}
\end{subfigure}
  \caption{(a) Data acquisition on the slice White-Male influencing the losses on the other slices for the UTKFace dataset. (b) To give an intuition, say there are three slices where shape indicates slice, and color indicates label. (Top) If we only increase the triangles, the decision boundary may shift to the left due to the new bias, changing the losses of the other slices. (Bottom) If we evenly increase the data for all slices, the bias does not change, and there is little influence among the slices.}
  \label{fig:influence}
\end{figure}

We now address the key challenge of handling unreliable learning curves. Learning curves are not perfect because slices may be too small for accurate estimations. Even worse, acquiring data for one slice may ``influence'' others. Figure~\ref{fig:utkfaceinfluence} shows how acquiring data for the slice White-Male increases or even decreases the model's loss on other slices for UTKFace. The intuition is that the acquired data of one slice pushes the decision boundary of the model, which in turn changes the losses of other slices (Figure~\ref{fig:influenceintuition}).


The solution is to iteratively update the learning curves. But how often should we iterate? On one hand, each iteration is expensive and involves multiple model trainings and curve fittings, even though we use amortization techniques~\cite{DBLP:journals/corr/abs-2003-04549}. On the other hand, we do not want to use inaccurate learning curves. Our algorithm works as follows. We first ensure a minimum slice size to draw some learning curve. In practice, having tens of examples is enough for this step. Next, we repeat two steps until we run out of budget: (1) acquire data as long as the estimated influence is not large enough and (2) re-fit the learning curves. The remaining problem is estimating influence. We propose a proxy called imbalance ratio change where imbalance ratio represents bias and is the ratio between the largest and smallest slice sizes. The intuition is that a change in imbalance ratio among slices causes influence. In Figure~\ref{fig:influenceintuition} adding two triangles results in a shifted decision boundary where the imbalance ratio increases from $\frac{2}{2} = 1$ to $\frac{4}{2} = 2$. On the other hand, if we evenly increase the slices, the decision boundary does not shift, and the imbalance ratio does not change much either.

In reference~\cite{DBLP:journals/corr/abs-2003-04549}, we provide more details on the algorithms and also perform experiments on real datasets. 
We show that \slicetuner{} has lower loss and unfairness compared to two baselines: uniformly acquiring the same amounts of data per slice  and waterfilling. We also make the same comparison when the slices are small and only have tens of examples. Here the learning curves are very noisy and thus unreliable. Interestingly, \slicetuner{} still outperforms the baselines because it can still leverage the relative loss differences among the learning curves. As more data is acquired, \slicetuner{} performs even better with more reliable learning curves. In the worst case when the learning curves are completely random, we expect \slicetuner{} to perform similarly to one of the baselines.

\subsection{Data Cleaning for Accurate, Fair, and Robust Models}

Another important place to support responsible AI is data cleaning~\cite{DBLP:books/acm/IlyasC19} where the input data needs to be validated and fixed before it is used for model training. Historically, multiple communities -- data management, machine learning (model fairness), and security -- have been investigating this problem under the names of data cleaning, unfairness mitigation, and data sanitization, respectively. Unfortunately, not much is known how the different techniques can be used together when a dataset is dirty, biased, and poisoned at the same time. 

\begin{figure}
\begin{floatrow}
\capbtabbox{%
    \begin{tabular}{| c | c | c | c | c | c |}
    \hline
    {\bf ID} & {\bf Weight} & {\bf Name} & {\bf Gender} & {\bf Age} & {\bf Label} \\\hline \hline
    $e_1$ & 1.0 & John & M & 20 & 1  \\\hline
    $e_2$ & 1.0 & Joe & M & 20 & 0 \\\hline
    $e_3$ & 1.0 & Joseph & M & 20 & 0 \\\hline
    $e_4$ & 1.0 & Sally & F & 30 & 1  \\\hline
    $e_5$ & 1.0 & Sally & F & 40 & 0  \\\hline
    $e_6$ & 1.0 & Sally & F & 300 & 1 \\\hline
    \end{tabular}
}{%
  \caption{Six examples where $e_2$ and $e_3$ are duplicates (dirty), and $e_6$ has an anomalous age (poisoned).}%
  \label{tbl:examples}
}
\ffigbox{%
  \includegraphics[scale=0.7]{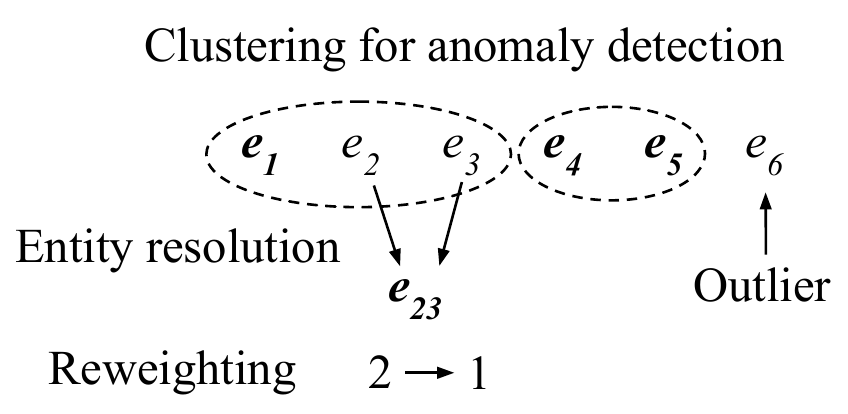}
}{%
  \caption{\mc{} running on the examples in Table~\ref{tbl:examples}.}%
  \label{fig:mlclean}
}
\end{floatrow}
\end{figure}

\mc{} is a unified cleaning framework that performs data cleaning, data sanitization, and unfairness mitigation together. A key insight is that these three operations have dependencies and must be executed in a certain order for the best performance. As shown in \mc{}'s architecture in Figure~\ref{fig:mlclean}, data sanitization and cleaning are performed together followed by unfairness mitigation. Data sanitization can be considered a stronger version of cleaning because it defends against adversarial poisoning instead of just noise. In addition, data cleaning and sanitization may affect the bias of data while unfairness mitigation that performs example reweighting does not affect the correctness of cleaning and sanitization. 

\begin{figure}[t]
  \centering
  \includegraphics[scale=0.7]{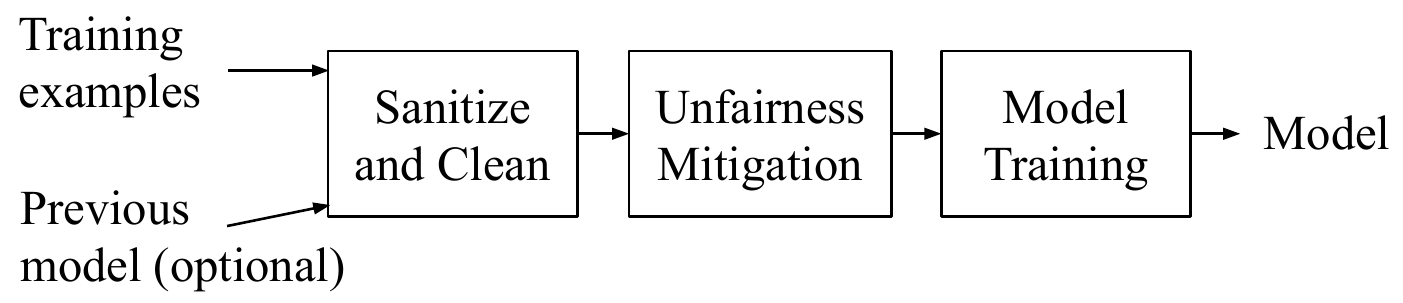}
  \caption{The \mc{} architecture where data sanitization and data cleaning are performed together followed by unfairness mitigation. Optionally, a previous model can be used by any of the three operations.}
  \label{fig:mlcleanarchitecture}
\end{figure}

As a running example, suppose we run \mc{} on the examples in Table~\ref{tbl:examples} with equal weights of 1. Say that data sanitization clusters examples and removes anomalies while data cleaning performs entity resolution. The two operations can be naturally combined by generating clusters and running entity resolution within each cluster, assuming that examples across clusters do not match. Clustering examples before resolution is a common operation in entity resolution for narrowing down matching candidates. Figure~\ref{fig:mlclean} shows how the initial six examples are clustered into \{$e_1$, $e_2$, $e_3$\} and \{$e_4$, $e_5$\} ($e_6$ is considered an outlier), and then $e_2$ and $e_3$ are merged together into $e_{23}$ with a summed weight of 2. For unfairness mitigation, suppose we reweight \cite{DBLP:journals/kais/KamiranC11} the examples such that demographic parity (defined in Section~\ref{sec:depth}) is satisfied for the sensitive groups men and women. We can make the (weighted) positive prediction rates the same by adjusting $e_{23}$'s weight from 2 to 1. As a result, the (weighted) positive prediction rates for men and women have the same value of  $\frac{1.0}{1.0 + 1.0} = 0.5$.

In reference~\cite{DBLP:conf/sigmod/TaeROKW19}, we compare \mc{} with other baselines that use a strict subset of the operations data sanitization, data cleaning, and unfairness mitigation or use all three operations, but in a different order than \mc{}. On real datasets, \mc{} has the best model accuracy and fairness, demonstrating that all three operations are necessary for the best results. In addition, \mc{} is faster than baselines that use the three operations in different orders, which means that utilizing the dependencies among the operations is important.

\section{Usable Responsible AI}
\label{sec:usability}

The final pillar of responsible AI is making it usable and actionable to all machine learning users. While usability is not always the main focus in machine learning, it is especially relevant for responsible AI because the various objectives are already conceptually challenging to understand, so the deployment must be made as easy as possible.
We thus propose two systems: \fb{}~\cite{fairbatch} is an easy-to-use model training technique for fairness, and \slicefinder{}~\cite{DBLP:conf/icde/ChungKPTW19,DBLP:journals/tkde/ChungKPTW20} is an easy-to-use model evaluation technique for improving fairness.

\subsection{Batch Selection for Fair Models}

While many unfairness mitigation techniques~\cite{aif360-oct-2018} have been proposed, most of them require significant amounts of effort to deploy. Pre-processing techniques have the advantage of being applicable to any model, but require changes in the training data in order to remove bias. In-processing techniques tend to perform well, but usually propose a new model training algorithm that completely replaces an existing algorithm. An interesting question is whether we can take the best of both worlds of pre-processing and in-processing without their overheads. 

\begin{figure}[t]
\centering
\begin{subfigure}{0.46\columnwidth}
\centering
\includegraphics[width=0.90\columnwidth,trim=0cm 0.25cm 0cm 0cm]{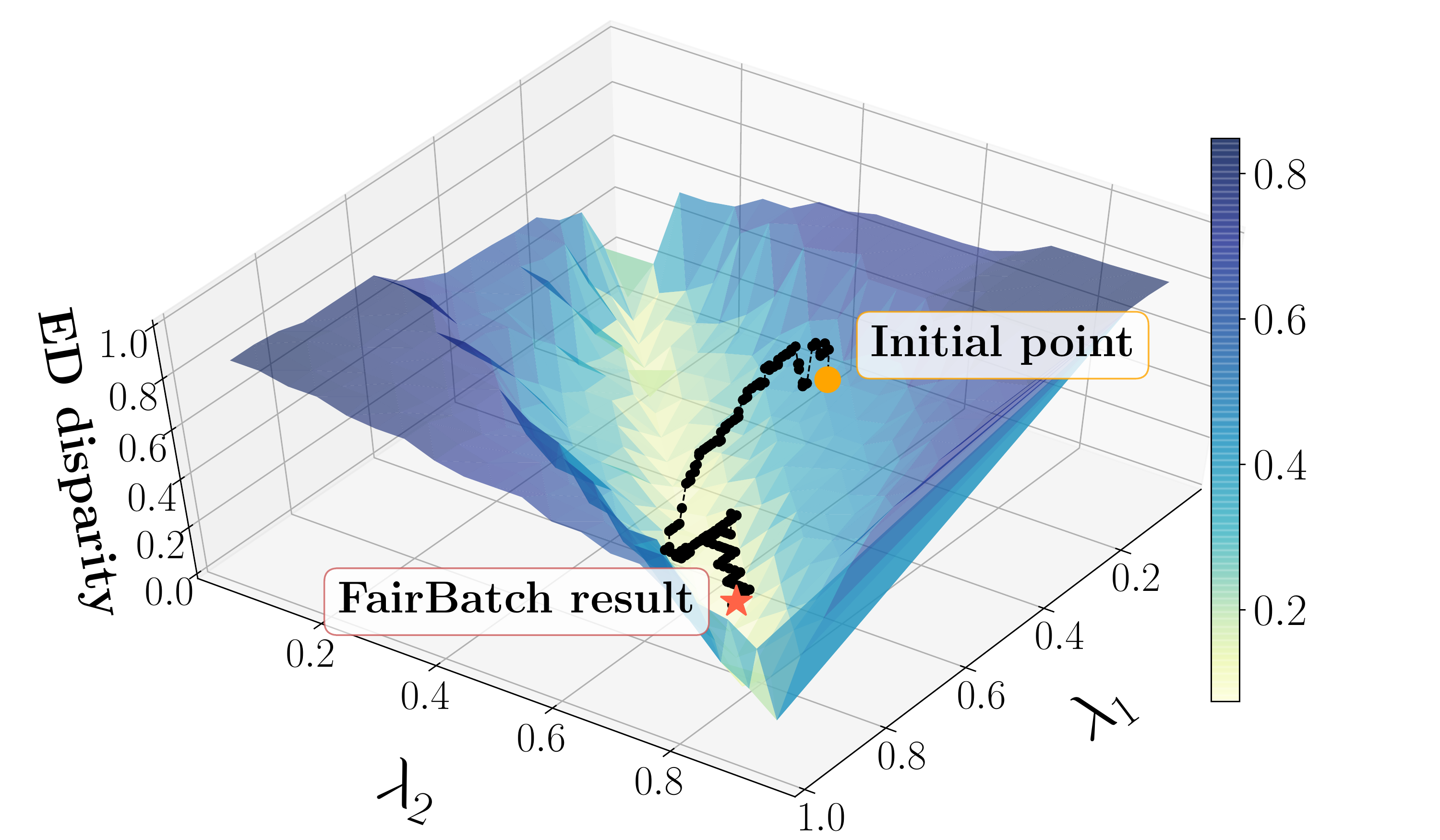}
\caption{}
\label{fig:gridsearch}
\end{subfigure}
\begin{subfigure}{0.51\columnwidth}
\centering
\includegraphics[scale=0.5]{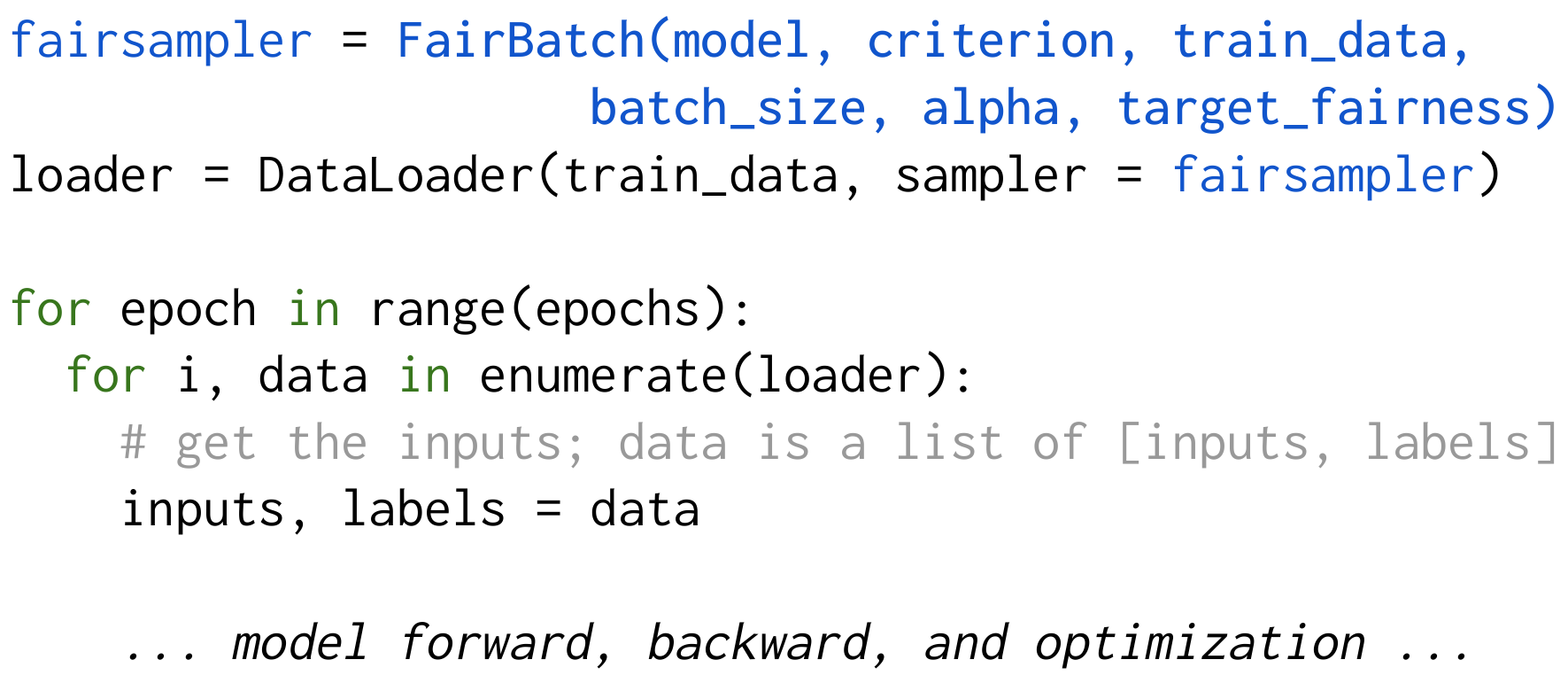}
\caption{}
\label{fig:pytorch_code}
\end{subfigure}
\caption{(a) The black path shows how the model fairness improves as \fb{} adjusts two parameters $\lambda_1$ (sampling rate for examples where $Z$=0 given $Y$=0) and $\lambda_2$ (sampling rate for examples where $Z$=0 given $Y$=1) for each epoch on the COMPAS dataset. ``ED disparity'' is the accuracy difference conditioned on the true label between sensitive groups where lower disparity means better equalized odds.
(b) Sample PyTorch code where the batch selection sampler is replaced with \fb{} with a single-line change highlighted in {\color{blue(ryb)} blue}.}
\end{figure}

We show that such a solution exists and propose \fb{}, which simply improves the batch selection of stochastic gradient descent training for better fairness. We formulate a bilevel optimization problem where we keep the standard training algorithm as the inner optimizer while incorporating the outer optimizer to equip the inner problem with the additional functionality: adaptively selecting minibatch sizes for the purpose of improving fairness. While the model is training, \fb{} adaptively adjusts the portions of the sensitive groups within each batch that is selected for each training epoch based on the fairness of the current intermediate model. For example, let us use the COMPAS example where we are predicting recidivism rates of criminals. Also let us use equalized odds (defined in Section~\ref{sec:depth}) as the fairness measure where we want the positive prediction rates of sensitive groups to be the same conditioned on the true label. Since the label is fixed, this fairness can be interpreted as the model having the same accuracy for sensitive groups conditioned on the label. Now suppose that an intermediate model shows higher accuracy for a certain sensitive group. \fb{} then increases the batch-size ratio of the other underperforming sensitive group in the next batch. Intuitively, a larger batch size ratio results in better accuracy, so eventually equalized odds will improve. Figure~\ref{fig:gridsearch} illustrates how \fb{} improves equalized odds during a single model training. In reference~\cite{fairbatch}, we show that this strategy is theoretically justified and generalize the algorithm for other fairness measures including demographic parity.



A key feature of \fb{} is its usability where one only needs to replace the batch selection of a machine learning system. Figure~\ref{fig:pytorch_code} shows a PyTorch code example where one can deploy \fb{} by replacing a single line of code, and no further changes are needed in the pre-processing or in-processing steps of model training.

In reference~\cite{fairbatch}, we also conduct experiments on synthetic and real datasets and show that \fb{} surprisingly has performances comparable to or even better than state-of-the-art pre-processing and in-processing unfairness mitigation techniques in terms of accuracy, fairness, and runtime. In addition, \fb{} is flexible and can be used to improve the fairness of pre-trained models like ResNet18 and GoogLeNet. Finally, there are batch selection techniques proposed for faster model training convergence, and \fb{} can be naturally combined with them to improve fairness as well.

\subsection{Automatic Data Slicing for Fair Models}

After model training, models are evaluated before being served. For example, TensorFlow Model Analysis~\cite{tfma} is a model evaluation component of TFX that accepts a user-specified slicing feature (e.g., country) and shows the model accuracies per slice (e.g., accuracy per country). Here we are using equalized error rates (defined in Section~\ref{sec:slicetuner}) as our notion of fairness.
However, there is potentially an exponential number of slices to explore, and it is not easy for users who do not have enough domain expertise to quickly sift through them. 

We thus propose \slicefinder{}~\cite{DBLP:conf/icde/ChungKPTW19,DBLP:journals/tkde/ChungKPTW20}, which automatically finds ``problematic'' slices (subsets of the data) where the model underperforms. Given these slices, users can take action by acquiring more data as in \slicetuner{} or debug the problematic data to find the root cause that led to the poor performance. We define a problematic slice to have the following characteristics. First, the slice must be interpretable where it can be defined with feature-value pairs, e.g., ``Gender=Male and Age=20-30.'' While one can also define a slice to be a cluster of examples, clusters are often difficult to understand in practice. In addition, the slice must have a relatively lower accuracy than its complement, i.e., the rest of the examples other than the slice, where the difference (effect size) is large and statistically significant. Finally, the slice must be large enough to have a meaningful impact on the overall model accuracy.

Since the search space for all possible slices is vast, we propose two approaches for searching. The first is a decision tree approach where we construct a decision tree of feature-value pairs to find slices. The traversal is fast, but the slices are non-overlapping, which means that we may miss some problematic slices. The second is a lattice search approach where we find slices by traversing a lattice of feature-value pairs in a breadth-first manner. Although we now find overlapping slices, this searching is slower than the decision tree approach. Once we find potential problematic slices, we perform effect-size and significance testings.

In references~\cite{DBLP:conf/icde/ChungKPTW19,DBLP:journals/tkde/ChungKPTW20}, we show that \slicefinder{} performs better than a clustering baseline on real datasets. Also while lattice searching is slower than decision tree searching, it finds more problematic slices.



\section{Open Challenges}
\label{sec:future}

We are far from achieving responsible AI for end-to-end machine learning and suggest promising directions. First, there needs to be deeper and broader support for the responsible AI objectives in each step of end-to-end machine learning. In addition, we believe the usability aspect of responsible AI has been largely understudied, and that there needs to be more emphasis on this important direction. Below are some concrete suggestions.

\begin{itemize}
\item {\em Data Collection}: We believe data acquisition must also support robustness. Dataset searching is becoming increasingly easy, and one challenge is distinguishing any poisoned data from the rest of the data. We also believe it is important to address fairness and robustness in data labeling. 

\item {\em Data Cleaning and Validation}: \mc{} is preliminary, and an interesting direction is to develop more general and automatic cleaning and validation techniques that support various combinations of data cleaning algorithms, fairness measures, and poisoning attacks.

\item {\em Model Training}: \frtrain{} is a first of its kind and can be extended in many ways. First, there needs to be more investigation on how to defend against more sophisticated poisoning attacks other than labeling flipping. Second, algorithm stability is a well-known issue in adversarial training and can be improved. Third, one may want to train models without a clean validation set.

\item {\em Model Evaluation}: There needs to be more robustness research for model evaluation where we can easily tell whether a model is accurate enough despite data poisoning in the training data.

\item {\em Model Management and Serving}: There needs to be more model managing and serving techniques that support fairness and robustness. While there are task-specific solutions like fairness in ranking~\cite{DBLP:journals/pvldb/AsudehJ20a}, an interesting direction is to generalize and support any task with minimal configuration.

\item There needs to be holistic solutions for the rest of the responsible AI objectives including explainability, transparency, and accountability. For example, recent data provenance and metadata~\cite{mlmd, bender-friedman-2018-data} solutions can be used to explain why each step in machine learning produced a certain result.

\end{itemize}





\section{Conclusion}

We proposed three research directions -- depth, breadth, and usability -- towards fully supporting responsible AI in end-to-end machine learning. While most research focuses on supporting one of many responsible AI features, we believe multiple objectives should be supported together, preferably in all steps from data collection to model serving. So far, we have scratched the surface of this vision where we proposed the following systems: \frtrain{} (holistic fair and robust training), \slicetuner{} (selective data acquisition for fair models), \mc{} (data cleaning for fair and robust models), \fb{} (easy-to-use batch selection for fair models), and \slicefinder{} (easy-to-use problematic slice finding for fair models). We also suggested various open challenges.

\section*{Acknowledgement}

This work was supported by a Google AI Focused Research Award and by the Engineering Research Center Program through the National Research Foundation of Korea (NRF) funded by the Korean Government MSIT (NRF-2018R1A5A1059921).


\end{document}